\documentclass[english,letterpaper, 10 pt, journal, twoside]{IEEEtran}
\usepackage[T1]{fontenc}
\usepackage[utf8]{inputenc}
\usepackage{color}
\usepackage{array}
\usepackage{refstyle}
\usepackage{booktabs}
\usepackage{url}
\usepackage{multirow}
\usepackage{amstext}
\usepackage{graphicx}
\usepackage{rotating}

\makeatletter

\AtBeginDocument{\providecommand\Tabref[1]{\ref{Tab:#1}}}
\AtBeginDocument{\providecommand\Figref[1]{\ref{Fig:#1}}}
\providecommand{\tabularnewline}{\\}
\RS@ifundefined{subsecref}
  {\newref{subsec}{name = \RSsectxt}}
  {}
\RS@ifundefined{thmref}
  {\def\RSthmtxt{theorem~}\newref{thm}{name = \RSthmtxt}}
  {}
\RS@ifundefined{lemref}
  {\def\RSlemtxt{lemma~}\newref{lem}{name = \RSlemtxt}}
  {}

\RequirePackage{todonotes}

\usepackage{amsmath,amsfonts,amssymb,mathrsfs}
\usepackage{algorithmic}
\usepackage{array}
\usepackage{stfloats}
\usepackage{url}
\usepackage{verbatim}
\usepackage{graphicx}
\usepackage{balance}

\usepackage{xcolor}
\usepackage{pifont}         
\usepackage{colortbl}
\usepackage{makecell}
\usepackage{stix}
\usepackage{animate}
\usepackage{threeparttable}
\usepackage{dirtree}
\usepackage{cite}
                               
\newcommand{\cmark}{\ding{51}}%
\newcommand{\xmark}{\ding{55}}%
\definecolor{alpha}{HTML}{FFCE19}
\definecolor{bob}{HTML}{FF00FF}
\definecolor{carol}{HTML}{19FFDC}
\definecolor{Gray}{gray}{0.9} 

\IEEEoverridecommandlockouts                              % This command is only needed if 
\let\NAT@parse\undefined
\usepackage[pagebackref,breaklinks,colorlinks,bookmarksopen,bookmarksnumbered,citecolor=blue,urlcolor=blue]{hyperref}

\@ifundefined{showcaptionsetup}{}{%
 \PassOptionsToPackage{caption=false}{subfig}}
\usepackage{subfig}
\makeatother

\usepackage{babel}
\begin{document}
\title{S3E: A Multi-Robot Multimodal Dataset for Collaborative SLAM}
\author{Dapeng Feng, Yuhua Qi, Shipeng Zhong, Zhiqiang Chen, \\
Qiming Chen, Hongbo Chen, Jin Wu, \IEEEmembership{Member, IEEE},
and Jun Ma\thanks{Manuscript received: September 3, 2024; Revised October 9, 2024; Accepted
October 24, 2024. This paper was recommended for publication by Editor
Javier Civera upon evaluation of the Associate Editor and Reviewers'
comments. Corresponding author: Yuhua Qi}\thanks{Dapeng Feng, Yuhua Qi, Shipeng Zhong, Zhiqiang Chen, and Hongbo Chen
are with Sun Yat-sen University, Guangzhou, China. {\tt\footnotesize \{fengdp5, zhongshp5, chenzhq56\}@mail2.sysu.edu.cn; \{qiyh8, chenhongbo\}@mail.sysu.edu.cn}}\thanks{Qiming Chen is with South China Agricultural University, Guangzhou,
China. {\tt\footnotesize chenqiming629@gmail.com}}\thanks{Jin Wu is with The Hong Kong University of Science and Technology,
Hong Kong SAR, China. {\tt\footnotesize jin\_wu\_uestc@hotmail.com}}\thanks{Jun Ma is with The Hong Kong University of Science and Technology
(Guangzhou), Guangzhou, China.{\tt\footnotesize jun.ma@ust.hk}}\thanks{Digital Object Identifier (DOI): see top of this page.}}
\maketitle
\begin{abstract}
The burgeoning demand for collaborative robotic systems to execute
complex tasks collectively has intensified the research community's
focus on advancing simultaneous localization and mapping (SLAM) in
a cooperative context. Despite this interest, the scalability and
diversity of existing datasets for collaborative trajectories remain
limited, especially in scenarios with constrained perspectives where
the generalization capabilities of Collaborative SLAM (C-SLAM) are
critical for the feasibility of multi-agent missions. Addressing this
gap, we introduce S3E, an expansive multimodal dataset. Captured by
a fleet of unmanned ground vehicles traversing four distinct collaborative
trajectory paradigms, S3E encompasses 13 outdoor and 5 indoor sequences.
These sequences feature meticulously synchronized and spatially calibrated
data streams, including 360-degree LiDAR point cloud, high-resolution
stereo imagery, high-frequency inertial measurement units (IMU), and
Ultra-wideband (UWB) relative observations. Our dataset not only surpasses
previous efforts in scale, scene diversity, and data intricacy but
also provides a thorough analysis and benchmarks for both collaborative
and individual SLAM methodologies. For access to the dataset and the
latest information, please visit our repository at \url{https://pengyu-team.github.io/S3E}. 

\end{abstract}


\section{Introduction}

\IEEEPARstart{C}{ollaborative} Simultaneous Localization and Mapping
(C-SLAM) is crucial for multi-agent cooperation, improving the robustness
and efficiency of localization and mapping in shared spaces \cite{Schmuck2021COVINS,Huang2022DiSCo-SLAM,zhong2023dcl,zhong2024colrio}.
Despite progress, repeatability and benchmarking in C-SLAM research
face challenges due to: 1) \textbf{\textit{System Complexity}}: The
complexity of C-SLAM systems, which integrate intricate software architectures
and a variety of hardware, complicates the process of achieving perfect
replication, thereby emphasizing the need for publicly available datasets
to facilitate methodological evaluation. 2) \textbf{\textit{Evaluation
Methodology}}: Traditional approaches, which divide single-agent SLAM
datasets into segments for virtual agents, are impractical for real-world
scenarios due to uniform perspectives and lighting at intersection
points \cite{Lajoie2021TCSLAM}. These limitations underscore the
importance of developing more effective evaluation strategies and
datasets to advance C-SLAM research.

Furthermore, effective trajectory design for multi-robot scenarios
in C-SLAM must follow two key principles: 1) \textbf{\textit{Temporal
and Spatial Diversity}}: Trajectories should be crafted to provide
multiple intra- and inter-loop closures at different times and places
along the trajectories, ensuring robots have complementary perspectives.
2) \textbf{\textit{Communication Constraints}}: Robots are typically
limited to sharing information within close proximity, necessitating
trajectory designs that maintain a reasonable interaction distance
for effective communication. In this paper, we introduce four trajectory
prototypes designed to meet these principles and evaluate the adaptability
of C-SLAM methodologies across diverse closure strategies in multi-robot
operations.

Publicly available datasets are crucial for the SLAM research community
for two main reasons: 1) \textbf{\textit{Development Acceleration}}:
Conducting specialized SLAM experiments involves significant investment
in advanced hardware and complex software, including calibration and
ground truth generation. Datasets aid in streamlining this development
process. 2) \textbf{\textit{Benchmarking and Analysis}}: Datasets
offer a standardized framework for comparing and assessing SLAM algorithms'
performance. While various SLAM datasets \cite{Geiger2013KITTI,Burri2016EuRoC}
have been introduced over the years, contributing to sensor diversity
and scenario breadth, most focus on single-agent systems. This focus
may limit their utility for multi-agent scenarios, highlighting the
need for more datasets that address multi-agent challenges.

The scarcity of datasets designed for C-SLAM is primarily due to the
increasing challenges in data acquisition with more collaborating
agents. Current C-SLAM datasets \cite{Leung2011UTIAS,Agarwal2020FordAV,Dubois2020AirMuseum,tian2023resilient,zhu2023graco,zhao2024subt}
often focus on a single cooperative approach with significant spatial
overlap, neglecting areas with minimal overlap. To fill this gap and
enhance C-SLAM research, we introduce S3E dataset, offering a multimodal
perspective with a variety of cooperative trajectory patterns in both
outdoor and indoor environments. 

Based on the S3E dataset, we conducted extensive experiments with
cutting-edge SLAM methods across various applications, from single-robot
to collaborative scenarios, using diverse sensor data. The findings
reveal that current C-SLAM techniques struggle with inter-loop closures
in some sequences, indicating a need for further research to enhance
the robustness and reliability of these systems.

In conclusion, our work makes several key contributions to the field:
\begin{itemize}
\item We have created a cutting-edge C-SLAM dataset using three ground robots,
each equipped with a 16-beam 3D laser scanner, two high-resolution
color cameras, a 9-axis IMU, a UWB receiver, and a dual-antenna RTK
receiver. This dataset is the first to incorporate UWB relative distance
measurements, providing a new research dimension.
\item To assess C-SLAM's performance in environments with limited overlap,
we have captured extensive long-term sequences using four meticulously
designed trajectory paradigms that address various intra- and inter-robot
loop closure scenarios.
\item We have conducted a thorough and multifaceted evaluation of current
C-SLAM techniques, employing both quantitative metrics and qualitative
analyses to assess their performance and utility.
\end{itemize}
\begin{table*}
\caption{\label{tab:dataset_comparison}Comparative Analysis of Prominent SLAM
Datasets. The abbreviations used within the table are as follows:
\textquotedbl Sw\textquotedbl{} denotes Software Synchronization,
indicating that synchronization across sensors or systems is achieved
through software mechanisms; \textquotedbl Hw\textquotedbl{} represents
Hardware Synchronization, where synchronization is managed by hardware
means. }

\centering
\resizebox{\linewidth}{!}{
\begin{centering}
{\scriptsize{}}%
\begin{tabular}{ll>{\columncolor{Gray}\centering}c>{\columncolor{Gray}\centering}c>{\columncolor{Gray}\centering}c>{\columncolor{Gray}\centering}ccc>{\columncolor{Gray}\centering}c>{\columncolor{Gray}\centering}cccl}
\toprule 
\multirow{2}{*}{Dataset} & \multirow{2}{*}{Platform} & \multicolumn{4}{>{\columncolor{Gray}\centering}c}{Sensors} & \multicolumn{2}{c}{Time Sync.} & \multicolumn{2}{>{\columncolor{Gray}\centering}c}{Trajectory} & \multicolumn{2}{c}{Environment} & \multirow{2}{*}{Ground Truth}\tabularnewline
 &  & Camera & IMU & LiDAR & UWB & Intra- & Inter- & Overlap & Paradigm & Indoor & Outdoor & \tabularnewline
\midrule 
\makecell[l]{KITTI \cite{Geiger2013KITTI}} & Car & \cmark & \cmark & \cmark & \xmark & Sw & \xmark & - & - & \xmark & \cmark & GNSS/INS\tabularnewline
\makecell[l]{EuRoC \cite{Burri2016EuRoC}} & UAV & \cmark & \cmark & \xmark & \xmark & Hw & \xmark & - & - & \cmark & \xmark & Motion Capture\tabularnewline
\makecell[l]{UTIAS \cite{Leung2011UTIAS}} & 5 UGVs & \cmark & \xmark & \xmark & \xmark & Sw & NTP & Large & 1 & \cmark & \xmark & Motion Capture\tabularnewline
\makecell[l]{AirMuseum \cite{Dubois2020AirMuseum}} & 3 UGVs, 1UAV & \cmark & \cmark & \xmark & \xmark & Sw & NTP & Large & 1 & \cmark & \xmark & SfM\tabularnewline
\makecell[l]{FordAV \cite{Agarwal2020FordAV}} & 3 Cars & \cmark & \cmark & \cmark & \xmark & - & GNSS & Large & 1 & \xmark & \cmark & GNSS/INS\tabularnewline
\makecell[l]{GRACO \cite{zhu2023graco}} & 1UGV, 1UAV & \cmark & \cmark & \cmark & \xmark & Hw & - & Restricted & 1 & \xmark & \cmark & GNSS/INS\tabularnewline
\makecell[l]{Kimera-Multi \cite{tian2023resilient}} & 8 UGVs & \cmark & \cmark & \cmark & \xmark & - & - & Restricted & 1 & \xmark & \cmark & \makecell[l]{GPS and total-station\\ assisted LiDAR SLAM}\tabularnewline
\makecell[l]{SubT-MRS \cite{zhao2024subt}} & \makecell[l]{UGVs, UAVs,\\ Handheld} & \cmark & \cmark & \cmark & \xmark & Hw & - & Restricted & 1 & \cmark & \cmark & 3D Scanner\tabularnewline
\midrule 
\textbf{Ours} & 3 UGVs & \cmark & \cmark & \cmark & \cmark & Hw & \makecell{GNSS,\\ PTPv2} & Restricted & 4 & \cmark & \cmark & \makecell[l]{RTK, GNSS/INS\\ Motion Capture}\tabularnewline
\bottomrule
\end{tabular}{\scriptsize\par}
\par\end{centering}
}

\vspace{-2mm}
\end{table*}

\begin{figure*}
\begin{centering}
\begin{minipage}[t]{0.6\linewidth}%
\begin{center}
\subfloat[\textbf{Sensor locations and the coordinate frames.}]{\begin{centering}
\par\end{centering}
\begin{centering}
\includegraphics[width=1\linewidth,height=4.5cm]{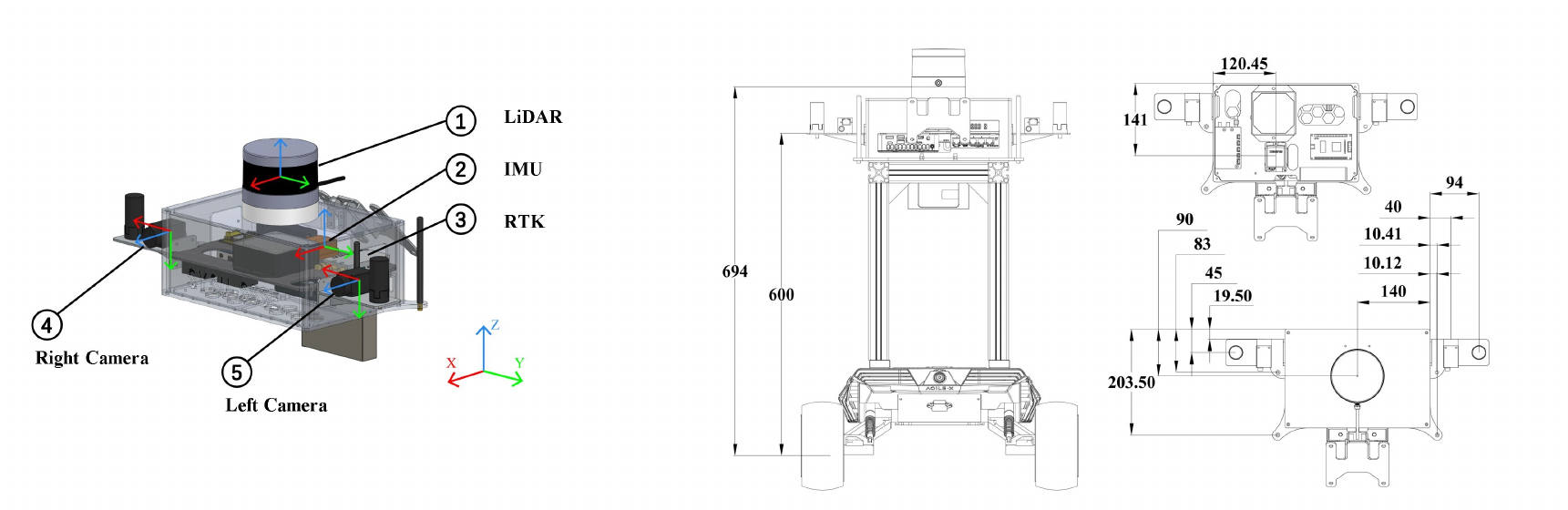}
\par\end{centering}
\centering{}}
\par\end{center}%
\end{minipage}\hfill{}%
\begin{minipage}[t]{0.4\linewidth}%
\begin{center}
\begin{minipage}[t]{0.5\linewidth}%
\begin{center}
\subfloat[\textbf{\label{fig:platform_v1.0}S3Ev1.0.}]{\begin{centering}
\par\end{centering}
\begin{centering}
\includegraphics[width=1\linewidth,height=4.5cm]{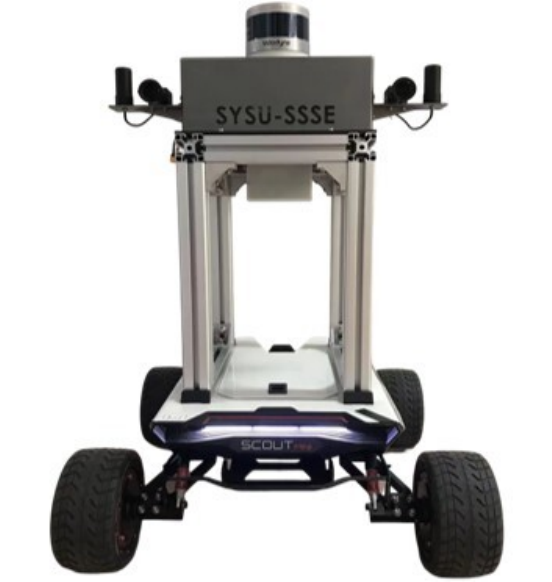}
\par\end{centering}
\centering{}}
\par\end{center}%
\end{minipage}%
\begin{minipage}[t]{0.5\linewidth}%
\begin{center}
\subfloat[\textbf{\label{fig:platform_v2.0}S3Ev2.0.}]{\begin{centering}
\par\end{centering}
\begin{centering}
\includegraphics[width=1\linewidth,height=4.5cm]{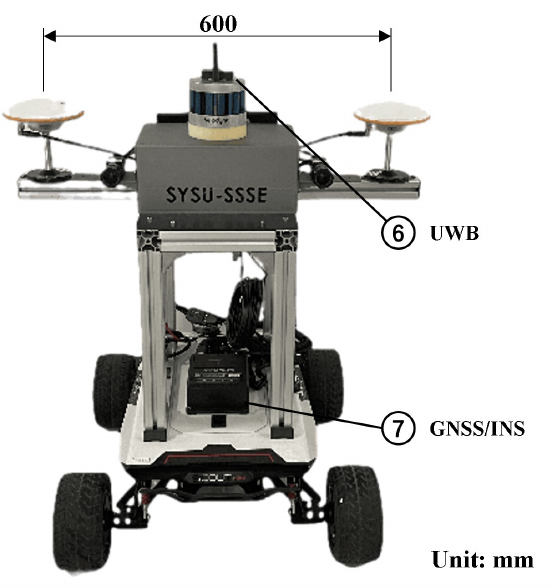}
\par\end{centering}
\centering{}}
\par\end{center}%
\end{minipage}
\par\end{center}%
\end{minipage}
\par\end{centering}
\caption{\label{fig:sensor_layout}\textbf{Mobile Platform Sensor Layout and
Coordinate Systems.} The left part details the sensor locations and
the coordinate frames that define their spatial orientation relative
to the platform. In the right part, our mobile platforms are available
in two versions, each designed for different operational requirements.}

\vspace{-6mm}
\end{figure*}

\section{Related Work}

\textbf{}

\textbf{Single-Agent SLAM Datasets}. High-quality, extensive datasets
are crucial for advancing SLAM research. The KITTI dataset \cite{Geiger2013KITTI},
known for its comprehensive urban data, has been a significant benchmark
for vision and LiDAR-based SLAM techniques. However, it has limitations,
such as the absence of perfectly synchronized data, offering only
estimated time delays between sensors. To address this, the EuRoC
dataset \cite{Burri2016EuRoC} was introduced, offering synchronized
visual and inertial data from a micro aerial vehicle in controlled
settings, using hardware clocks for synchronization.

\textbf{Multi-Agent SLAM Datasets.}  The UTIAS dataset \cite{Leung2011UTIAS}
was a pioneering work with five robots exploring a small area. The
AirMuseum dataset \cite{Dubois2020AirMuseum} introduced a heterogeneous
multi-robot setup in a warehouse, and the FordAV dataset \cite{Agarwal2020FordAV}
is collected by a fleet of autonomous vehicles, providing a multi-agent
perspective with seasonal variations in weather, lighting, construction,
and traffic conditions. However, these datasets often have significant
co-view overlap, limiting comprehensive C-SLAM evaluation.

The GRACO dataset \cite{zhu2023graco} attempted to address this by
introducing a ground-aerial approach, but it had synchronization issues
and did not fully handle dynamic environments. The SubT-MRS dataset
\cite{zhao2024subt} added more variety with multi-robot SLAM data
across different environments. Kimera-Multi \cite{tian2023resilient}
advanced the state with a comprehensive collection featuring diverse
trajectory designs and multiple robots.

It's important to note that existing datasets, including GRACO, SubT-MRS,
and Kimera-Multi, often overlook the complexities of collaborative
trajectory patterns. The ability to handle intra- and inter-loop closures
is crucial for C-SLAM robustness and is not consistently addressed.
Considering these collaborative patterns is essential for developing
C-SLAM algorithms that can operate effectively in environments with
varying agent interactions and dynamic scenes.

In this paper, we introduce a large-scale C-SLAM dataset that includes
visual ambiguities and dynamic objects, captured under four distinct
trajectory paradigms to assess C-SLAM generalizability under restricted
overlap conditions. Our dataset also incorporates UWB relative measurement
positioning, adding a significant dimension for C-SLAM research. For
a comprehensive comparison of different datasets, including our contribution,
refer to \Tabref{dataset_comparison}.

\section{S3E Dataset}

 The S3E dataset is meticulously assembled with a focus on high temporal
precision in sensor data synchronization. Each sensor is calibrated
to a shared timescale for harmonized multi-sensory information, facilitated
by an advanced time synchronization mechanism. This is crucial for
accurate data integration across different modalities. The dataset
features two mobile robot platform versions:
\begin{itemize}
\item S3Ev1.0: Designed for indoor use with a compact design for exceptional
maneuverability in tight spaces.
\item S3Ev2.0: Enhanced with a wider frame to accommodate a high-precision
ground truth system and an UWB module for improved localization accuracy
in challenging environments.
\end{itemize}
For a visual representation of sensor placement, refer to \Figref{sensor_layout}.

\subsection{Sensor Configuration}

Our S3E dataset encompasses a multimodal array of sensors, each selected
for its operational range and noise characteristics, and meticulously
synchronized to capture data with high temporal precision. The sensors
are integrated onto the \textit{Agilex Scout Mini}, a versatile all-terrain
mobile platform capable of high-speed remote-controlled navigation,
featuring four-wheel drive and a maximum speed of $10km/h$.

 The technical specifications of the sensors and ground truth devices
integrated into our payload are outlined on our project website\footnote{\url{https://pengyu-team.github.io/S3E}}.
This includes the sensor types, their resolution, measurement range,
accuracy, and any other pertinent technical details that define their
contribution to the SLAM system's performance.\textcolor{red}{{} }In
particular, we utilize the centermeter-accuracy GNSS and RTK for outdoor
environments, achieving a precision of $\pm1\text{cm}$, to record
the ground truth trajectories. For indoor environments, where GNSS
signals are typically unavailable, we utilize a millimeter-precision
motion capture system to obtain our ground truth data.

\subsection{Sensor Synchronization}

This section delves into the critical processes of time synchronization
and sensor calibration, which are essential for achieving optimal
sensor fusion and maximizing system performance within a dynamic,
multimodal environment.

\subsubsection{Time Synchronization}

Time synchronization across all sensors is essential for the accurate
co-registration of multi-sensor data, particularly in dynamic environments
where precise timing is critical to the quality of the fused dataset.
This synchronization ensures that data from various sensors is temporally
aligned, facilitating coherent integration and analysis.

Our synchronization system is built around an \textit{Altera EP4CE10}
FPGA board acting as the primary trigger device, with an \textit{Intel
NUC11TNKv7} serving as the host computer. The system is designed with
a comprehensive set of I/O interfaces to accommodate the diverse requirements
of different sensors. For synchronization across agents, we address
two distinct scenarios:
\begin{itemize}
\item In outdoor settings with access to Global Navigation Satellite System
(GNSS) signals, we use GNSS time as the global reference to synchronize
the timing across agents. 
\item In GNSS-denied environments such as indoors and tunnels, agents synchronize
their timers by obtaining external global time data from Alpha, which
serves as a Precision Time Protocol version 2 (PTPv2) server, via
a wireless network connection.
\end{itemize}

Considering transmission delays, all sensor readings are forwarded
to the host computer, where they are timestamped upon arrival, organized,
and packaged to ensure accurate temporal referencing for subsequent
analysis.

\subsubsection{Sensor Calibration}

Our sensor suite is governed by a unified coordinate system framework,
adhering to the right-hand rule, which ensures uniformity in data
orientation and facilitates standardized analysis.

The calibration process was meticulously executed through five distinct
stages: camera-intrinsic, IMU-intrinsic, camera-camera extrinsic,
camera-IMU extrinsic, and camera-lidar extrinsic. Each stage was crucial
for ensuring the accurate alignment and integration of sensor data.

Upon completion of these calibration stages, the resulting parameters
were meticulously stored in YAML files for each agent, ensuring that
the calibrated parameters are readily accessible and reproducible.

\subsection{Ground Truth}

To create accurate ground truth tracks for our dataset, we use a three-pronged
approach:

For outdoor environments with good GNSS signal reception, a dual-antenna
RTK device is used to achieve highly accurate localization data with
centimeter-level precision. The S3Ev2.0 platform features a GNSS/INS
unit that significantly enhances ground truth instrumentation. This
integration allows for high-frequency positioning data output even
during GNSS signal outages, such as in tunnels. The system ensures
continuous and reliable ground truth data by fusing RTK measurements
with IMU data using post-processing fusion techniques.

In indoor environments without GNSS signals, like laboratories, a
motion capture system with 17 high-frequency cameras is used to record
track start and endpoints due to the high costs of a large-scale system.
In other indoor areas lacking a motion capture system, such as teaching
buildings, the RTK device is used to record the initial and terminal
points of the tracks outside the buildings.

By combining these methods, we are able to generate comprehensive
ground truth positions that cater to both outdoor and indoor environments,
thereby facilitating a robust evaluation of C-SLAM algorithms across
diverse real-world scenarios.

\begin{figure}
\begin{centering}
\par\end{centering}
\begin{centering}
\includegraphics[width=1\linewidth]{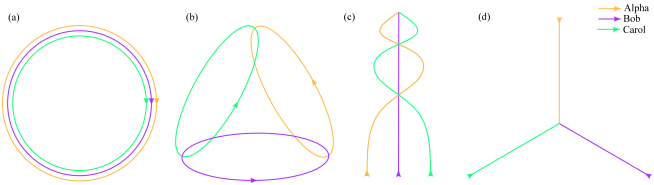}
\par\end{centering}
\caption{\label{fig:trajecotry_paradigms}\textbf{Illustration of Trajectory
Paradigms for C-SLAM. }The distinct trajectory paradigms adopted by
three agents, designated as Alpha, Bob, and Carol, to demonstrate
different interaction and information exchange patterns in a multi-agent
SLAM context. \textbf{(a)} \textbf{Concentric Circles. (b) Intersecting
Circles. (c) Intersection Curve. (d) Rays }}

\vspace{-2mm}
\end{figure}

\subsection{\label{subsec:Trajectories}Trajectory Paradigms}

The S3E dataset incorporates a diverse set of trajectory paradigms
to simulate various collaborative scenarios that agents may encounter
during SLAM operations. As despited in \Figref{trajecotry_paradigms},
we have designed these paradigms to address different loop closure
conditions, both within individual agents (intra-loop) and across
multiple agents (inter-loop). The following are the four trajectory
paradigms we have included in our dataset:

\textbf{Concentric Circles:} This paradigm involves coordinated unmanned
platforms that stay connected and collect similar environmental data.
This allows for efficient area coverage and tasks like environmental
monitoring and mapping. Real-time data sharing improves accuracy,
but the similarity of observations can make it challenging to detect
and adapt to rapid environmental changes.

\textbf{Intersecting Circles:} In this paradigm, unmanned platforms
work separately in different areas and only share data when they intersect.
This approach reduces the need for constant communication and coordination
but may limit the consistency of overall mapping and positioning due
to less frequent data exchange.

\begin{table}
\caption{\textbf{\label{tab:Trajectory-Paradigms.}}Comparative Analysis of
Trajectory Paradigms for Loop Closure Detection in C-SLAM. The symbols
within the table indicate the presence ($\mdlgblkcircle$) or absence
($\mdlgwhtcircle$) of loop closures for each paradigm, providing
a quick reference for the suitability of each strategy in different
collaborative robotic missions.}

\resizebox{\linewidth}{!}{
\begin{centering}
\begin{tabular}{lcc>{\raggedright}m{0.5\linewidth}}
\toprule 
\multirow{2}{*}[-1mm]{Trajectory Paradigm} & \multicolumn{2}{c}{Closures} & \multirow{2}{0.5\linewidth}[-1mm]{Applications}\tabularnewline
\cmidrule{2-3} \cmidrule{3-3} 
 & Intra-loop & Inter-loop & \tabularnewline
\midrule 
\rowcolor{Gray}Concentric Circles & $\blackcircleulquadwhite$ & $\mdlgblkcircle$ & Extensive coverage and detailed exploration\tabularnewline
Intersecting Circles & $\mdlgblkcircle$ & $\blackcircleulquadwhite$ & Distributed search and rescue operations\tabularnewline
\rowcolor{Gray}Intersection Curve & $\mdlgwhtcircle$ & $\circlerighthalfblack$ & Large-scale distributed exploration, patrol, and mapping\tabularnewline
Rays & $\mdlgwhtcircle$ & $\circleurquadblack$ & Independent exploration and mapping\tabularnewline
\bottomrule
\end{tabular}
\par\end{centering}
}

\vspace{-6mm}
\end{table}

\textbf{Intersection Curve:} This paradigm involves unmanned platforms
starting from different locations, meeting at specific points, and
eventually converging at a final destination. They communicate and
share data only during these meetings, which helps in large-scale
exploration and mapping by reducing cumulative errors. Each platform's
independent operation boosts efficiency and coverage. However, this
approach requires advanced navigation for planning effective meeting
points and routes.

\textbf{Rays:} In this paradigm, autonomous platforms begin at various
starting points, work independently, and come together at a common
endpoint for communication and data exchange. This method is effective
for large-scale, independent operations like disaster recovery and
mining exploration, where each platform operates on its own. The final
convergence helps in sharing historical data and optimizing local
maps, but the limited interaction can affect the overall consistency
of mapping and positioning, thus requiring advanced autonomous navigation
and decision-making skills.

Each paradigm is meticulously crafted to offer a robust framework
for assessing C-SLAM algorithms across a multitude of real-world collaborative
robotic applications. \Tabref{Trajectory-Paradigms.} provides a comparative
analysis of the loop closure detection situations corresponding to
these paradigms, highlighting their respective advantages and potential
application areas.

\begin{table*}[t]
\caption{\label{tab:Analysis-of-S3E}Analysis of S3E dataset. }

\centering
\resizebox{\linewidth}{!}{
\begin{centering}
\begin{tabular}{cclcccccccccccccc}
\toprule 
\multicolumn{3}{c}{Scenario} & \multirow{2}{*}{Time{[}s{]}} & \multicolumn{4}{c}{Trajectory} & \multirow{2}{*}{Ground Truth} & \multicolumn{3}{c}{Length{[}m{]}} & \multirow{2}{*}{Size{[}GB{]}} & \multicolumn{4}{c}{Sensors}\tabularnewline
\cmidrule{1-3} \cmidrule{2-3} \cmidrule{3-3} \cmidrule{5-8} \cmidrule{6-8} \cmidrule{7-8} \cmidrule{8-8} \cmidrule{10-12} \cmidrule{11-12} \cmidrule{12-12} \cmidrule{14-17} \cmidrule{15-17} \cmidrule{16-17} \cmidrule{17-17} 
\multicolumn{2}{c}{Env.} & Region &  & a & b & c & d &  & Alpha ($\alpha$) & Bob ($\beta$) & Carol ($\gamma$) &  & Camera & IMU & LiDAR & UWB\tabularnewline
\midrule
\multirow{13}{*}{\begin{turn}{90}
Outdoor
\end{turn}} & \multirow{7}{*}{\begin{turn}{90}
v1.0
\end{turn}} & Square\_1 & 460 &  &  & \cmark &  & RTK & 546.0 & 496.5 & 529.2 & 17.8 & \cmark & \cmark & \cmark & \xmark\tabularnewline
 &  & Square\_2 & 255 &  &  &  & \cmark & RTK & - & 250.6 & 246.4 & 9.4 & \cmark & \cmark & \cmark & \xmark\tabularnewline
 &  & Library\_1 & 454 & \cmark &  &  &  & RTK & 507.6 & 517.2 & 498.9 & 16.3 & \cmark & \cmark & \cmark & \xmark\tabularnewline
 &  & Campus\_Road\_1 & 878 &  & \cmark &  &  & RTK & 920.5 & 995.9 & 1072.3 & 29.4 & \cmark & \cmark & \cmark & \xmark\tabularnewline
 &  & Playground\_1 & 298 & \cmark &  &  &  & RTK & 407.7 & 425.6 & 445.5 & 8.7 & \cmark & \cmark & \cmark & \xmark\tabularnewline
 &  & Playground\_2 & 222 &  &  & \cmark &  & RTK & 265.6 & 315.7 & 456.4 & 6.3 & \cmark & \cmark & \cmark & \xmark\tabularnewline
 &  & Dormitory\_1 & 671 &  &  &  & \cmark & RTK & 727.0 & 719.3 & 721.9 & 23.5 & \cmark & \cmark & \cmark & \xmark\tabularnewline
\cmidrule{2-17} \cmidrule{3-17} \cmidrule{4-17} \cmidrule{5-17} \cmidrule{6-17} \cmidrule{7-17} \cmidrule{8-17} \cmidrule{9-17} \cmidrule{10-17} \cmidrule{11-17} \cmidrule{12-17} \cmidrule{13-17} \cmidrule{14-17} \cmidrule{15-17} \cmidrule{16-17} \cmidrule{17-17} 
 & \multirow{6}{*}{\begin{turn}{90}
v2.0
\end{turn}} & Square\_3 & 466 &  &  & \cmark &  & GNSS/INS & 487.4 & 569.6 & 563.8 & 8.5 & \xmark & \cmark & \cmark & \cmark\tabularnewline
 &  & Library\_2 & 491 & \cmark &  &  &  & GNSS/INS & 521.6 & 523.9 & 520.3 & 9.3 & \xmark & \cmark & \cmark & \cmark\tabularnewline
 &  & Campus\_Road\_2 & 1594 & \cmark &  &  &  & GNSS/INS & 1938.6 & 1934.2 & 1950.1 & 30.9 & \xmark & \cmark & \cmark & \cmark\tabularnewline
 &  & Campus\_Road\_3 & 907 & \cmark &  &  &  & GNSS/INS & 983.3 & 967.6 & 986.9 & 17.5 & \xmark & \cmark & \cmark & \cmark\tabularnewline
 &  & Playground\_3 & 111 &  & \cmark &  &  & GNSS/INS & 84.7 & 91.3 & 110.7 & 1.8 & \xmark & \cmark & \cmark & \cmark\tabularnewline
 &  & Tunnel\_1 & 425 & \cmark &  &  &  & GNSS/INS & 521.9 & 502.4 & 501.1 & 8.2 & \xmark & \cmark & \cmark & \cmark\tabularnewline
\midrule 
\multirow{5}{*}{\begin{turn}{90}
Indoor
\end{turn}} & \multirow{5}{*}{\begin{turn}{90}
v1.0
\end{turn}} & Teaching\_Building\_1 & 798 &  & \cmark &  &  & RTK & 617.2 & 734.4 & 643.4 & 27.3 & \cmark & \cmark & \cmark & \xmark\tabularnewline
 &  & Laboratory\_1 & 292 &  & \cmark &  &  & Motion Capture & 147.7 & 161.5 & 141.0 & 9.6 & \cmark & \cmark & \cmark & \xmark\tabularnewline
 &  & Laboratory\_2 & 391 &  & \cmark &  &  & Motion Capture & 215.3 & 199.1 & 160.3 & 12.7 & \cmark & \cmark & \cmark & \xmark\tabularnewline
 &  & Laboratory\_3 & 410 &  & \cmark &  &  & Motion Capture & 219.1 & 202.2 & 204.2 & 13.3 & \cmark & \cmark & \cmark & \xmark\tabularnewline
 &  & Laboratory\_4 & 380 & \cmark &  &  &  & Motion Capture & 173.7 & 177.2 & 180.0 & 12.7 & \cmark & \cmark & \cmark & \xmark\tabularnewline
\bottomrule
\end{tabular}
\par\end{centering}
}

\vspace{-2mm}
\end{table*}

\begin{figure*}[t]
\begin{centering}
\par\end{centering}
\begin{centering}
\includegraphics[width=0.88\linewidth]{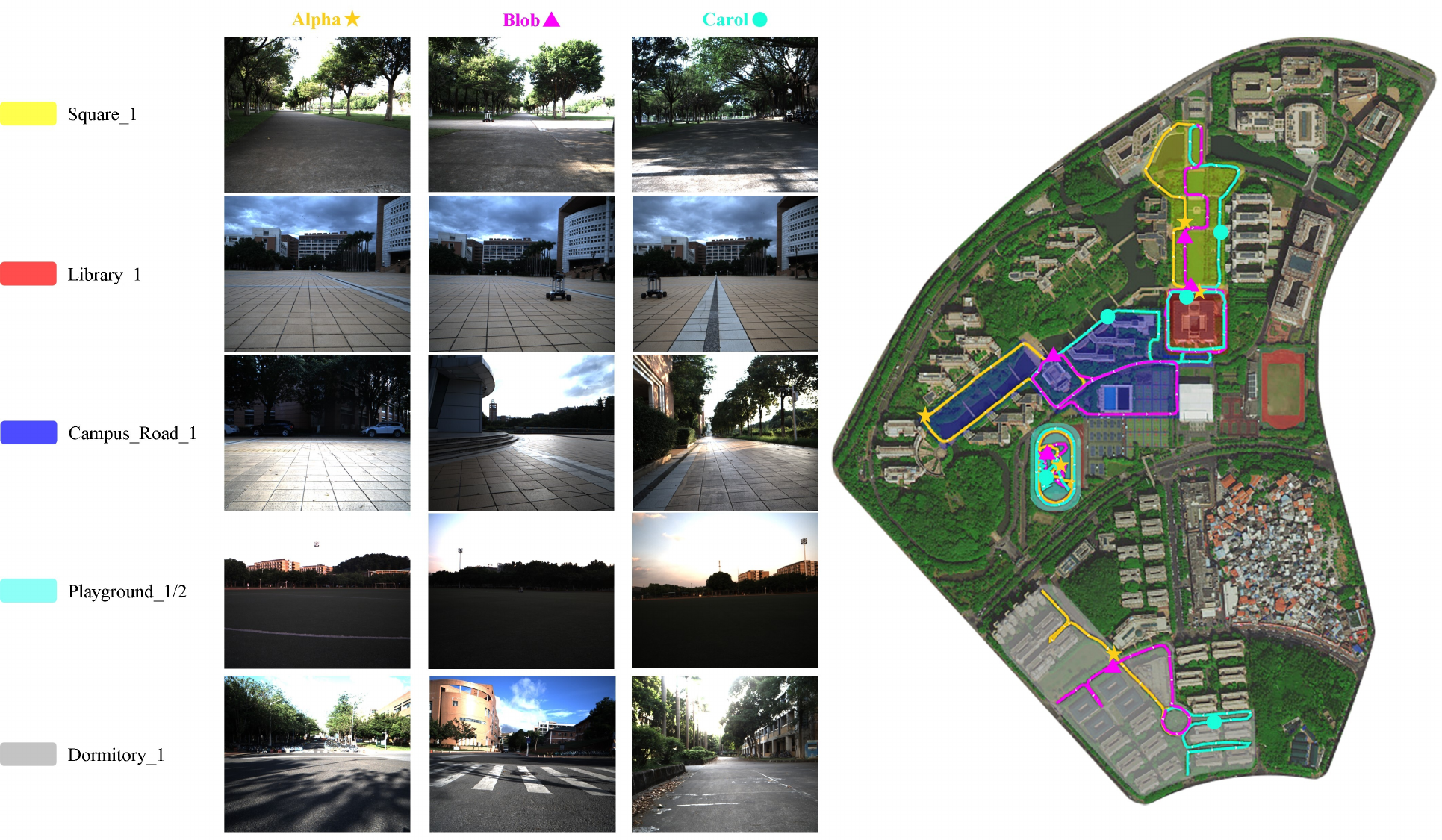}
\par\end{centering}
\caption{\textbf{\label{fig:outdoor_tracks}Visualization of Outdoor Trajectory
in the S3E Dataset.} The outdoor trajectories captured in the S3E
dataset by three tele-operated mobile platforms, designated as Alpha
($\alpha$), Bob ($\beta$), and Carol ($\gamma$). The trajectories
are distinctly annotated with \textcolor{alpha}{Orange}, \textcolor{bob}{Purple}, and \textcolor{carol}{Cyan}.
The annotations \textcolor{alpha}{$\bigstar$}, \textcolor{bob}{$\blacktriangle$}, and \textcolor{carol}{$\mdblkcircle$}
indicate the specific positions where data was recorded in each sequence.
The left portion of the figure presents the synchronized data capture
at the annotated points, demonstrating the collaborative data collection
process. }

\vspace{-6mm}
\end{figure*}

\begin{figure}
\newcommand{\database}{ \raisebox{-0.25\width}{\includegraphics[height=0.5cm]{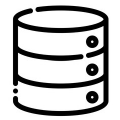}} }
\newcommand{\folder}{ \raisebox{-0.25\width}{\includegraphics[height=0.5cm]{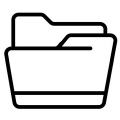}} }
\newcommand{\bin}{ \raisebox{-0.25\width}{\includegraphics[height=0.5cm]{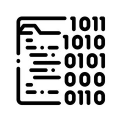}} }
\newcommand{\txt}{ \raisebox{-0.25\width}{\includegraphics[height=0.5cm]{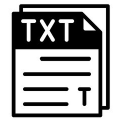}} }
\newcommand{\yaml}{ \raisebox{-0.25\width}{\includegraphics[height=0.5cm]{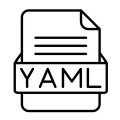}} }
\dirtree{%
 .1 \database S3E.
 .2 \folder <\textbf{platform\_id}>.
 .3 \folder Calibration.
 .4 \yaml <\textbf{agent\_id}>.yaml.
 .3 \folder S3E\_<\textbf{sequence\_id}>.  
 .4 \bin S3E\_<\textbf{sequence\_id}>.db3.  
 .4 \yaml metadata.yaml.  
 .4 \txt <\textbf{agent\_id}>\_gt.txt.
} 

\caption{\textbf{\label{fig:data_organization}S3E Dataset Organizational Structure.}}

\vspace{-10mm}
\end{figure}

\subsection{Dataset Analysis}

To test collaborative mission scenarios and trajectory paradigms,
we conducted data collection at Sun Yat-sen University's Guangzhou
East Campus using three tele-operated robots under human control for
safety. The dataset, detailed in \Tabref{Analysis-of-S3E}, includes
diverse environments like squares and libraries, with 13 outdoor and
5 indoor sequences. It totals over 263.2 GB, 475.2 minutes of footage,
and covers more than 28.1 km. With an average sequence length of 527.9
seconds, it's suitable for long-term C-SLAM technology evaluations. 

To thoroughly assess the accuracy and robustness of C-SLAM algorithms
in complex, real-world scenarios, our dataset encompasses a diverse
range of environments, each presenting unique challenges:

\textbf{Dormitory}: High pedestrian and bicycle traffic test perception
and tracking. Regular layouts are ideal for assessing C-SLAM precision
and consistency across platforms.

\textbf{Campus Road}: Long distances and expansive views evaluate
endurance and large-scale exploration. Long-distance and multi-cycle
data assess stability, accuracy, and efficiency over extended operations.

\textbf{Playground}: Open spaces with fewer obstructions challenge
feature extraction and optimization. Data collected at different times
of day test adaptability to lighting conditions and performance under
rapid motion.

\textbf{Laboratory}: Confined spaces with complex layouts and rich
semantic content test advanced scene understanding and mapping capabilities.

\textbf{Teaching Building and Tunnel}: Poor lighting and similar geometric
structures challenge robustness in maintaining accurate positioning
and mapping, especially in tunnels and corridors.

Our dataset features unique areas like libraries and squares, adding
environmental variety and perceptual challenges. \Figref{outdoor_tracks}
shows the trajectory paradigms in outdoor settings within the S3E
dataset. The dataset covers a range of challenging environments that
C-SLAM algorithms may encounter, including dynamic objects, long operation
times, perceptual aliasing, indoor settings, and significant motion.
This diversity is crucial for evaluating C-SLAM performance, adaptability,
and robustness, which are key for advancing collaborative robotic
navigation and mapping technologies.

\begin{table*}[t]
\caption{\label{tab:message}Descriptions of ROS2 topics included in our dataset.
Note that \textbf{agent\_id} stands for Alpha, Blob, and Carol.}

\begin{threeparttable}
\centering
\resizebox{\linewidth}{!}{
\begin{centering}
\begin{tabular}{lll>{\raggedright}m{0.3\linewidth}l}
\toprule 
Sensor & Topic & Type & Description & Frequency\tabularnewline
\midrule 
LiDAR & /\textbf{agent\_id}/velodyne\_points & sensor\_msgs/msg/PointCloud2 & Raw velodyne pointcloud with $[x,y,z,intensity,ring,time]$ & 10 $Hz$\tabularnewline
\midrule
\multirow{2}{*}[-3mm]{Camera\footnotemark[1]} & /\textbf{agent\_id}/left\_camera/compressed & sensor\_msgs/msg/CompressedImage & Compressed RGB8 image with $1224\times1024$ & \multirow{2}{*}[-3mm]{10 $Hz$}\tabularnewline
\cmidrule{2-4} \cmidrule{3-4} \cmidrule{4-4} 
 & /\textbf{agent\_id}/right\_camera/compressed & sensor\_msgs/msg/CompressedImage & Compressed RGB8 image with $1224\times1024$ & \tabularnewline
\midrule
IMU & /\textbf{agent\_id}/imu/data & sensor\_msgs/msg/Imu & Raw imu data & 100 $Hz$\tabularnewline
\midrule
RTK\footnotemark[2] & /\textbf{agent\_id}/fix & sensor\_msgs/msg/NavSatFix & RTK-GNSS data with $[latitude,longitude,altitude]$ & 1 $Hz$\tabularnewline
\midrule
UWB\footnotemark[3] & /\textbf{agent\_id}/nlink\_linktrack\_nodeframe2 & std\_msgs/msg/Float64MultiArray & UWB data comprises 13 data fields \footnotemark[4] & 50 $Hz$\tabularnewline
\midrule
\multirow{5}{*}[-10mm]{GNSS/INS\footnotemark[3]} & /\textbf{agent\_id}/time\_reference & sensor\_msgs/msg/TimeReference & GPS time reference anchored at the epoch of 1980-01-06 00:00:00 UTC & \multirow{5}{*}[-10mm]{100 $Hz$}\tabularnewline
\cmidrule{2-4} \cmidrule{3-4} \cmidrule{4-4} 
 & /\textbf{agent\_id}/fix & sensor\_msgs/msg/NavSatFix & RTK-GNSS data with $[latitude,longitude,altitude]$ & \tabularnewline
\cmidrule{2-4} \cmidrule{3-4} \cmidrule{4-4} 
 & /\textbf{agent\_id}/correct\_imu & sensor\_msgs/msg/Imu & Raw build-in imu data & \tabularnewline
\cmidrule{2-4} \cmidrule{3-4} \cmidrule{4-4} 
 & /\textbf{agent\_id}/vel & geometry\_msgs/msg/TwistStamped & Timestamped linear velocities $(m/s)$ of mobile platform along $[east,north,up]$ & \tabularnewline
\cmidrule{2-4} \cmidrule{3-4} \cmidrule{4-4} 
 & /\textbf{agent\_id}/heading & geometry\_msgs/msg/QuaternionStamped & Timestamped quaternions map mobile platform attitudes & \tabularnewline
\bottomrule
\end{tabular}
\par\end{centering}
}
\begin{tablenotes}
\footnotesize
\item \footnotemark[1] Available in all sequences of S3Ev1.0. \footnotemark[2] Available in outdoor sequences of S3Ev1.0. \footnotemark[3] Available in all sequences of S3Ev2.0. 
\item \footnotemark[4] Detailed definitions of UWB data fields are documented and accessible on our project website.
\end{tablenotes}
\end{threeparttable}

\vspace{-2mm}
\end{table*}

\subsection{Dataset Format}

\begin{table*}
\caption{\label{tab:baseline_1} ATE $[m]$ for Single SLAM and C-SLAM in the
S3Ev1.0 outdoor environment without UWB measurement.{\scriptsize{}
}\xmark~fails to initialize or track frames. If inter-loop closures
detection fails, we mark it ``Failed''.}

\centering
\resizebox{\linewidth}{!}{
\begin{centering}
\begin{tabular}{l>{\columncolor{Gray}\centering}c>{\columncolor{Gray}\centering}c>{\columncolor{Gray}\centering}cccc>{\columncolor{Gray}\centering}c>{\columncolor{Gray}\centering}c>{\columncolor{Gray}\centering}cccc>{\columncolor{Gray}\centering}c>{\columncolor{Gray}\centering}c>{\columncolor{Gray}\centering}cccc>{\columncolor{Gray}\centering}c>{\columncolor{Gray}\centering}c>{\columncolor{Gray}\centering}c}
\toprule 
\multirow{2}{*}{Methods} & \multicolumn{3}{>{\columncolor{Gray}\centering}c}{Square\_1} & \multicolumn{3}{c}{Square\_2} & \multicolumn{3}{>{\columncolor{Gray}\centering}c}{Library\_1} & \multicolumn{3}{c}{Campus\_Road\_1} & \multicolumn{3}{>{\columncolor{Gray}\centering}c}{Playground\_1} & \multicolumn{3}{c}{Playground\_2} & \multicolumn{3}{>{\columncolor{Gray}\centering}c}{Dormitory\_1}\tabularnewline
 & $\alpha$ & $\beta$ & $\gamma$ & $\alpha$ & $\beta$ & $\gamma$ & $\alpha$ & $\beta$ & $\gamma$ & $\alpha$ & $\beta$ & $\gamma$ & $\alpha$ & $\beta$ & $\gamma$ & $\alpha$ & $\beta$ & $\gamma$ & $\alpha$ & $\beta$ & $\gamma$\tabularnewline
\midrule
ORB-SLAM3 & 1.16 & 15.5 & \xmark & - & 2.81 & \xmark & 11.4 & \xmark & \xmark & \xmark & 55.5 & \xmark & 0.87 & \xmark & \xmark & 3.29 & \xmark & \xmark & \multicolumn{3}{>{\columncolor{Gray}\centering}c}{\xmark}\tabularnewline
VINS-Fusion & 1.81 & \xmark & 4.83 & - & 1.51 & 0.62 & 7.95 & 7.86 & 5.56 & \xmark & 16.5 & \xmark & 3.24 & 7.36 & 4.31 & 31.3 & \xmark & \xmark & 6.67 & 3.97 & 7.70\tabularnewline
LIO-SAM & 1.19 & 1.75 & \xmark & - & 0.73 & 0.36 & 1.12 & 1.52 & 1.14 & 2.06 & 3.25 & 2.43 & \xmark & \xmark & 0.86 & \xmark & \xmark & 0.68 & 0.63 & 1.44 & 0.91\tabularnewline
LVI-SAM & 1.21 & 0.88 & \xmark & - & 0.79 & 0.40 & 1.89 & 1.67 & 1.31 & 2.44 & 3.14 & 1.30 & \xmark & 1.59 & 0.76 & 6.78 & 6.10 & 0.72 & 0.86 & 1.48 & 0.94\tabularnewline
\makecell[l]{CoLRIO\\ (front-end)} & \textbf{0.90} & 0.57 & 0.65 & - & \textbf{0.49} & \textbf{0.19} & 1.37 & 1.54 & \textbf{0.88} & \textbf{1.06} & 0.91 & 1.10 & 0.39 & 0.45 & 0.20 & 0.69 & 0.35 & 0.29 & \textbf{0.47} & \textbf{0.83} & 0.81\tabularnewline
\midrule 
COVINS & 1.63 & 0.83 & 1.75 & \multicolumn{3}{c}{Failed} & \multicolumn{3}{>{\columncolor{Gray}\centering}c}{\xmark} & 18.3 & 9.61 & 54.7 & \multicolumn{3}{>{\columncolor{Gray}\centering}c}{\xmark} & \multicolumn{3}{c}{\xmark} & \multicolumn{3}{>{\columncolor{Gray}\centering}c}{\xmark}\tabularnewline
DiSCo-SLAM & \multicolumn{3}{>{\columncolor{Gray}\centering}c}{Failed} & \multicolumn{3}{c}{Failed} & 0.74 & 1.33 & 1.27 & 2.21 & 1.35 & Failed & \textbf{0.27} & \textbf{0.37} & 0.38 & \multicolumn{3}{c}{\xmark} & 0.54 & 1.47 & Failed\tabularnewline
Swarm-SLAM & 8.13 & 7.11 & 2.13 & - & 1.39 & 0.52 & 3.16 & 2.71 & 3.22 & 7.62 & 11.3 & 6.36 & 3.36 & 3.38 & 1.70 & 1.36 & 1.32 & 2.05 & 10.0 & 5.61 & 4.50\tabularnewline
DCL-SLAM & 1.23 & 0.86 & \textbf{0.66} & - & 0.50 & 0.26 & \textbf{0.58} & 1.26 & 1.17 & 1.51 & 1.48 & 1.87 & 0.33 & 0.40 & 0.32 & \multicolumn{3}{c}{\xmark} & 0.52 & 1.37 & \textbf{0.51}\tabularnewline
CoLRIO & 0.91 & \textbf{0.53} & 0.80 & - & 0.56 & 0.20 & 1.06 & \textbf{1.05} & 1.02 & 1.08 & \textbf{0.79} & \textbf{1.08} & 0.39 & 0.60 & \textbf{0.16} & \textbf{0.44} & \textbf{0.32} & \textbf{0.29} & 0.54 & 0.87 & 0.84\tabularnewline
\bottomrule
\end{tabular}
\par\end{centering}
}

\vspace{-4mm}
\end{table*}

\begin{table}
\caption{\label{tab:baseline_2}ATE $[m]$ for Single SLAM and C-SLAM in the
S3Ev1.0 indoor environment without UWB measurement.}

\resizebox{\linewidth}{!}{
\begin{centering}
\begin{tabular}{l>{\columncolor{Gray}\centering}c>{\columncolor{Gray}\centering}c>{\columncolor{Gray}\centering}cccc>{\columncolor{Gray}\centering}c>{\columncolor{Gray}\centering}c>{\columncolor{Gray}\centering}cccc}
\toprule 
\multirow{2}{*}{Methods} & \multicolumn{3}{>{\columncolor{Gray}\centering}c}{Laboratory\_1} & \multicolumn{3}{c}{Laboratory\_2} & \multicolumn{3}{>{\columncolor{Gray}\centering}c}{Laboratory\_3} & \multicolumn{3}{c}{Laboratory\_4}\tabularnewline
 & $\alpha$ & $\beta$ & $\gamma$ & $\alpha$ & $\beta$ & $\gamma$ & $\alpha$ & $\beta$ & $\gamma$ & $\alpha$ & $\beta$ & $\gamma$\tabularnewline
\midrule 
LIO-SAM & 0.47 & 1.82 & 2.51 & 12.2 & 6.12 & 8.88 & 6.24 & 8.57 & 7.31 & 1.96 & 0.88 & 0.65\tabularnewline
FAST-LIO2 & 7.27 & 11.2 & 5.13 & 10.2 & 2.14 & 15.0 & 12.8 & 6.80 & 7.33 & 0.79 & 0.88 & 0.37\tabularnewline
\makecell[l]{CoLRIO\\ (front-end)} & \textbf{0.40} & 0.51 & 0.44 & 0.83 & 0.79 & 0.96 & 1.79 & 0.91 & 2.43 & \textbf{0.67} & 0.65 & 0.87\tabularnewline
\midrule 
DiSCo-SLAM & 0.56 & \textbf{0.43} & \textbf{0.33} & 2.02 & 0.36 & 4.60 & 2.01 & \textbf{0.31} & 1.28 & 1.40 & 2.02 & 3.45\tabularnewline
Swarm-SLAM & 9.86 & 1.14 & 3.29 & 4.56 & 7.03 & 15.3 & 7.06 & 12.2 & 8.10 & 1.79 & 3.40 & 3.47\tabularnewline
CoLRIO & 0.43 & 0.50 & 0.44 & \textbf{0.40} & \textbf{0.07} & \textbf{0.08} & \textbf{0.49} & 0.91 & \textbf{0.13} & 0.95 & \textbf{0.65} & \textbf{0.18}\tabularnewline
\bottomrule
\end{tabular}
\par\end{centering}
}

\vspace{-6mm}
\end{table}

\begin{table*}
\caption{\label{tab:baseline_3} ATE $[m]$ for C-SLAM in the S3Ev2.0 outdoor
environment with UWB measurement.}

\centering
\resizebox{\linewidth}{!}{
\begin{centering}
{\scriptsize{}}%
\begin{tabular}{l>{\columncolor{Gray}\centering}c>{\columncolor{Gray}\centering}c>{\columncolor{Gray}\centering}cccc>{\columncolor{Gray}\centering}c>{\columncolor{Gray}\centering}c>{\columncolor{Gray}\centering}cccc>{\columncolor{Gray}\centering}c>{\columncolor{Gray}\centering}c>{\columncolor{Gray}\centering}cccc}
\toprule 
\multirow{2}{*}{Methods} & \multicolumn{3}{>{\columncolor{Gray}\centering}c}{Square\_3} & \multicolumn{3}{c}{Library\_2} & \multicolumn{3}{>{\columncolor{Gray}\centering}c}{Campus\_Road\_2} & \multicolumn{3}{c}{Campus\_Road\_3} & \multicolumn{3}{>{\columncolor{Gray}\centering}c}{Playground\_3} & \multicolumn{3}{c}{Tunnel\_1}\tabularnewline
 & $\alpha$ & $\beta$ & $\gamma$ & $\alpha$ & $\beta$ & $\gamma$ & $\alpha$ & $\beta$ & $\gamma$ & $\alpha$ & $\beta$ & $\gamma$ & $\alpha$ & $\beta$ & $\gamma$ & $\alpha$ & $\beta$ & $\gamma$\tabularnewline
\midrule
DiSCo-SLAM & \textbf{1.15} & \textbf{1.26} & 12.11 & 1.07 & 1.02 & 0.98 & 15.41 & 14.60 & 3.37 & 1.27 & 1.32 & 39.76 & 9.01 & 0.72 & 2.32 & 5.06 & 11.90 & 4.40\tabularnewline
Swarm-SLAM & 8.52 & 5.64 & 2.80 & 3.15 & 0.91 & 0.94 & 48.96 & 64.23 & 22.37 & 8.79 & 9.60 & 4.90 & 1.11 & 3.24 & 5.47 & 10.49 & 11.06 & 10.06\tabularnewline
\makecell[l]{CoLRIO\\ (w/o UWB)} & 1.79 & 1.62 & \textbf{1.30} & 0.91 & 0.89 & 0.88 & \textbf{1.29} & 2.97 & \textbf{1.21} & 1.04 & 1.13 & 1.07 & 0.24 & \textbf{0.19} & 0.30 & 5.49 & 4.69 & 5.15\tabularnewline
\midrule
\makecell[l]{CoLRIO\\ (w UWB)} & 1.71 & 1.52 & 1.54 & \textbf{0.89} & \textbf{0.87} & \textbf{0.88} & 1.37 & \textbf{1.59} & 1.34 & \textbf{0.77} & \textbf{0.61} & \textbf{0.90} & \textbf{0.23} & 0.20 & \textbf{0.30} & \textbf{4.01} & \textbf{3.21} & \textbf{3.60}\tabularnewline
\bottomrule
\end{tabular}{\scriptsize\par}
\par\end{centering}
}

\vspace{-4mm}
\end{table*}

Our research utilizes the ROS2 \cite{macenski2022robot} bag format
for sensor data storage, a standard in robotics known for efficient
data management and playback. To simplify data replay and ensure easy
access to synchronized sensor streams from our Alpha ($\alpha$),
Blob ($\beta$), and Carol ($\gamma$) robots, we've implemented a
strategy to merge data from the same operational sequence into one
ROS2 bag file. 

\textbf{Data Organization}: Our dataset, as shown in \Figref{data_organization}
, is meticulously organized with a clear directory structure. Calibration
parameters are detailed in YAML files within a specific Calibration
folder. Each data sequence is complete with a \texttt{.db3} file for
primary sensor measurements and a \texttt{metadata.yaml} file outlining
sequence specifics like sensor configurations and capture conditions.
Additionally, to aid in thorough performance assessments, we've included
auxiliary files like \texttt{<}\texttt{\textbf{agent\_id}}\texttt{>\_gt.txt},
which contain ground truth data.

\textbf{Ground Truth Format}: The ground truth data, which is essential
for evaluating the accuracy of C-SLAM algorithms, is provided as TXT
files. These files contain timestamped poses in UTM coordinates and
orientation quaternions, formatted as follows: $\left[timestamp,t_{x},t_{y},t_{z},q_{x},q_{y},q_{z},q_{w}\right]$.
The position $t_{a\in\{x,y,z\}}$ denotes the robot's location in
UTM coordinates, while $q_{a\in\{x,y,z,w\}}$ represents the orientation
in quaternion form.

\textbf{Sensor Data Topics}: \Tabref{message} outlines the sensor
topics included in the dataset, detailing the type of data each topic
contains, its description, and the frequency of data capture. 

\section{Experiments}

\subsection{Baselines}

We have implemented four single-agent SLAM systems, namely ORB-SLAM3
\cite{Campos2021ORB-SLAM3}, VINS-Fusion \cite{Qin2019VINS-Fusion},
LIO-SAM \cite{Shan2020LIO-SAM}, and LVI-SAM \cite{Shan2021LVI-SAM}.
Additionally, we have incorporated five collaborative SLAM (C-SLAM)
systems into our study, which include COVINS \cite{Schmuck2021COVINS},
DiSCo-SLAM \cite{Huang2022DiSCo-SLAM}, Swarm-SLAM \cite{Lajoie2023SwarmSLAM},
DCL-SLAM \cite{zhong2023dcl}, and CoLRIO \cite{zhong2024colrio}.
These systems have been evaluated using the S3E dataset.

For most of the baselines, we only modify the intrinsic and extrinsic
of the sensors and use the left camera for evaluation.

\begin{figure}
\begin{centering}
\par\end{centering}
\begin{centering}
\includegraphics[width=0.85\linewidth]{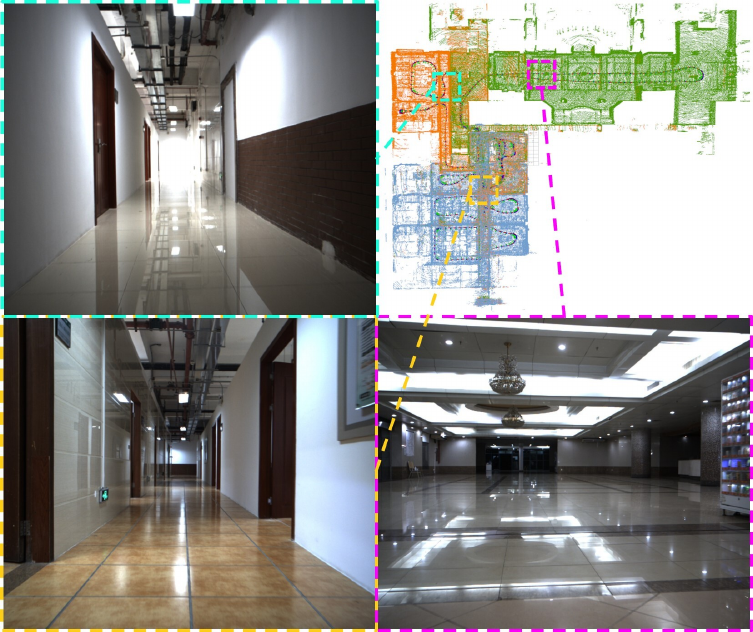}
\par\end{centering}
\caption{\textbf{\label{fig:qualitative_result_2}Map in Laboratory\_1 with
CoRLIO.}}

\vspace{-6mm}
\end{figure}

\begin{figure*}
\begin{centering}
\begin{minipage}[t]{0.33\linewidth}%
\subfloat[\textbf{Tunnel\_1 with DiSCo-SLAM.}]{\begin{centering}
\par\end{centering}
\begin{centering}
\includegraphics[width=1\linewidth,height=5cm]{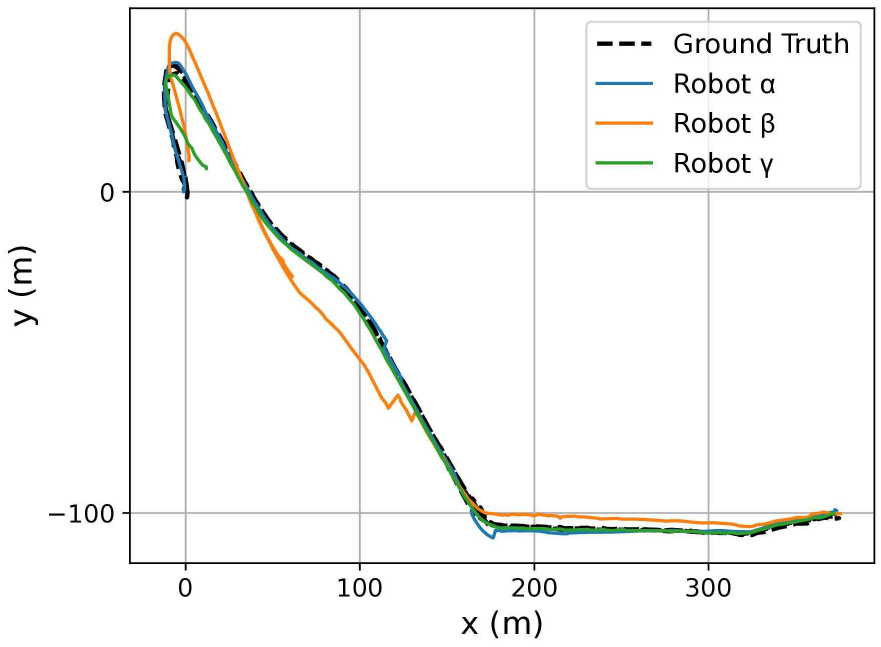}
\par\end{centering}
}%
\end{minipage}\hfill{}%
\begin{minipage}[t]{0.33\linewidth}%
\subfloat[\textbf{Tunnel\_1 with Swarm-SLAM.}]{\begin{centering}
\par\end{centering}
\begin{centering}
\includegraphics[width=1\linewidth,height=5cm]{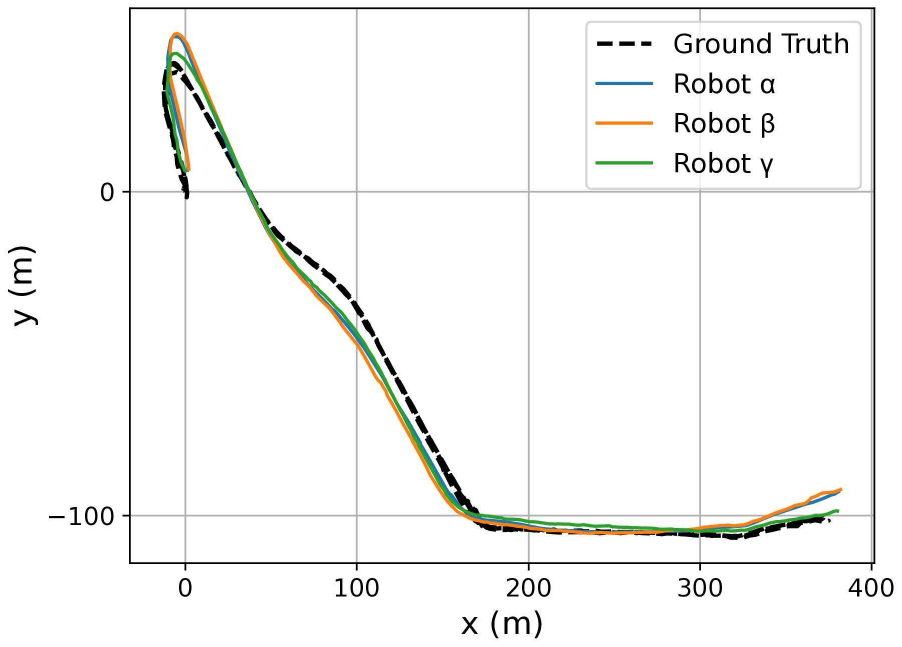}
\par\end{centering}
}%
\end{minipage}\hfill{}%
\begin{minipage}[t]{0.33\linewidth}%
\subfloat[\textbf{Tunnel\_1 with CoLRIO.}]{\begin{centering}
\par\end{centering}
\begin{centering}
\includegraphics[width=1\linewidth,height=5cm]{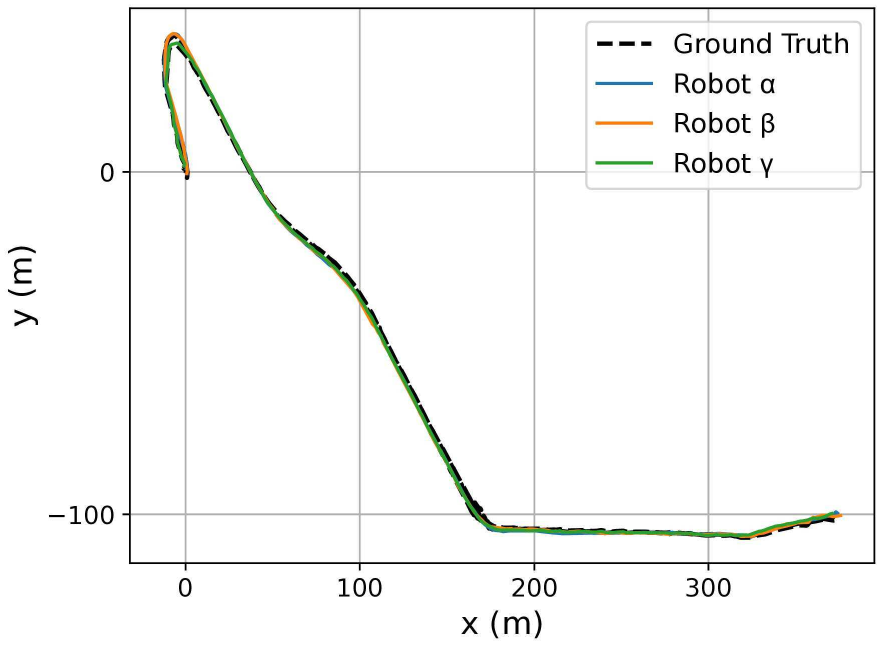}
\par\end{centering}
}%
\end{minipage}
\par\end{centering}
\begin{centering}
\begin{minipage}[t]{0.33\linewidth}%
\subfloat[\textbf{Campus\_Road\_2 with DiSCo-SLAM.}]{\begin{centering}
\par\end{centering}
\begin{centering}
\includegraphics[width=1\linewidth,height=5cm]{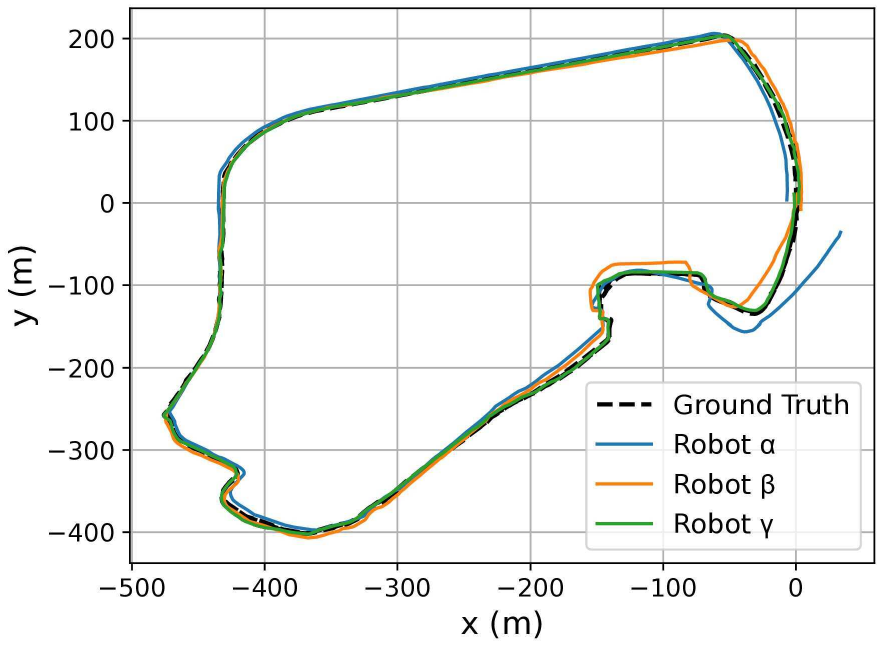}
\par\end{centering}
}%
\end{minipage}\hfill{}%
\begin{minipage}[t]{0.33\linewidth}%
\subfloat[\textbf{Campus\_Road\_2 with Swarm-SLAM.}]{\begin{centering}
\par\end{centering}
\begin{centering}
\includegraphics[width=1\linewidth,height=5cm]{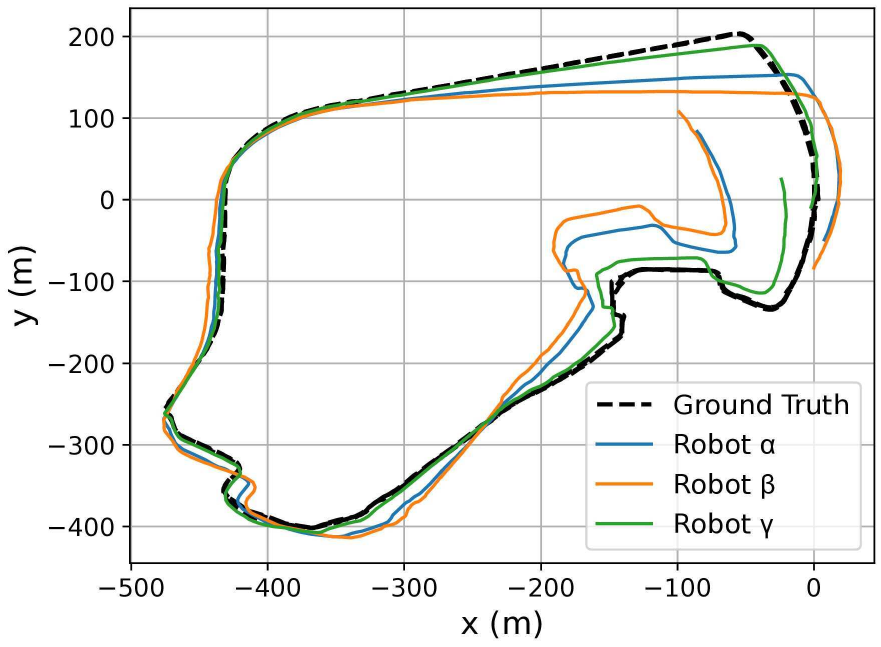}
\par\end{centering}
}%
\end{minipage}\hfill{}%
\begin{minipage}[t]{0.33\linewidth}%
\subfloat[\textbf{Campus\_Road\_2 with CoLRIO.}]{\begin{centering}
\par\end{centering}
\begin{centering}
\includegraphics[width=1\linewidth,height=5cm]{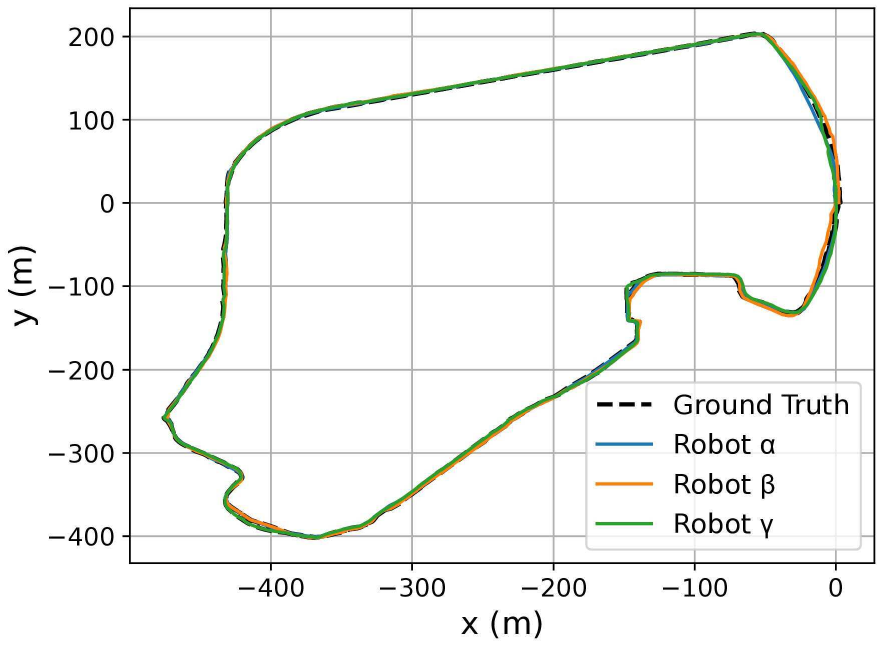}
\par\end{centering}
}%
\end{minipage}
\par\end{centering}
\caption{\textbf{\label{fig:qualitative_results_3}The qualitative results
of outdoor environments.}}

\vspace{-4mm}
\end{figure*}

\subsection{Results}

Our experimental evaluation of the S3E dataset provides valuable insights
into the performance of various state-of-the-art SLAM methodologies
under diverse real-world conditions. The results, summarized in \Tabref{baseline_1}
and \Tabref{baseline_2} , reveal the absolute trajectory error (ATE)
for both single-agent and collaborative SLAM (C-SLAM) systems in outdoor
and indoor environments without UWB measurement.
\begin{itemize}
\item \textbf{Key Findings:}
\begin{itemize}
\item \textbf{Inter-Loop Closures Detection Impact}: The results highlight
the increased performance of C-SLAM systems when successful in detecting
inter-loop closures, showcasing the benefits of collaborative information
sharing among agents.
\item \textbf{Front-end State Estimation Impact}: The performance of C-SLAM
is heavily influenced by the accuracy of its front-end state estimation.
For example, CoLRIO's success in certain scenarios is attributed to
its effective front-end tracker.
\end{itemize}
\item \textbf{Specific Observations}:
\begin{itemize}
\item In scenarios with large spatial overlap, C-SLAM systems leveraged
inter-robot measurements to enhance state estimation accuracy. However,
in areas with limited overlap, reducing drift remained a challenge.
\item The incorporation of UWB measurements in CoLRIO significantly improved
localization robustness and accuracy, as demonstrated in \Tabref{baseline_3},
showcasing the benefits of additional ranging data for C-SLAM systems.
\end{itemize}
\end{itemize}
The results from our experiments on the S3E dataset underscore the
importance of continuous innovation in C-SLAM algorithms. The varying
performance across different scenarios and the impact of collaborative
data sharing highlight the need for further research. The qualitative
results presented in  \Figref{qualitative_result_2} and \Figref{qualitative_results_3},
visually complement the performance metrics, providing a deeper understanding
of the C-SLAM systems' behavior in diverse environments.

\section{Known Issues}

The S3E dataset, while a substantial contribution to the field of
C-SLAM, presents certain limitations. This section delves into the
known issues associated with the dataset and the broader challenges
of C-SLAM systems:
\begin{enumerate}
\item Scalability: C-SLAM systems must efficiently handle operations with
a large number of robots without significant performance degradation.
\item Resource Management: There are strict limitations on communication
and computation capabilities of mobile robots, which C-SLAM must overcome
to achieve real-time performance.
\end{enumerate}

\section{Conclusion}

The S3E dataset provides a comprehensive and multifaceted platform
for evaluating C-SLAM systems under a variety of real-world conditions.
It features four distinct trajectory paradigms for broad scenario
evaluation. Captured across diverse indoor and outdoor environments,
the dataset includes meticulously synchronized sensor data, offering
high detail and complexity. Our experiments using this dataset have
highlighted the improved robustness of C-SLAM systems, especially
in handling inter-loop closures. The addition of UWB measurements
has significantly enhanced localization accuracy and reliability.
The S3E dataset establishes a new benchmark for C-SLAM, fostering
further research and innovation in multi-agent robotic navigation
and mapping, and advancing the field.

\section*{Acknowledgments}

\noindent This research greatly benefited from the guidance and expertise
of many contributors. We extend our profound gratitude to our colleagues,
Yizhen Yin and Haoxin Zhang from Sun Yat-sen University, for significantly
improving our dataset's quality and applicability, especially in data
collection. Special acknowledgment goes to Prof. Tao Jiang and his
student Yudu Jiao from Chongqing University for evaluating the single-agent
SLAM algorithms on our dataset.

\section*{}

\bibliographystyle{IEEEtran}
\bibliography{bibtex/slam,bibtex/dataset,bibtex/3d-construction,bibtex/simulator,bibtex/synchronization}

\end{document}